# K+ Means : An Enhancement Over K-Means Clustering Algorithm

*Srikanta Kolay*
*SMS India Pvt. Ltd., RDB Boulevard*
*5th Floor, Unit-D, Plot No.-K1, Block-EP&GP,*
*Sector-V, Salt Lake, Kolkata-700091, India*
*Email: kolaysrikanta@gmail.com*

*Kumar S. Ray*
*Electronics and Communication Science Unit*
*Indian Statistical Institute*
*203, B.T Road, Kolkata-700108, India*
*E-mail: ksray@isical.ac.in*

*Abhoy Chand Mondal*
*Department of Computer Science*
*University of Burdwan*
*Burdwan - 713104*
*West Bengal, India*
*E-mail: abhoy_mondal@yahoo.co.in*

**Abstract**
K-means (MacQueen, 1967) [1] is one of the simplest unsupervised learning algorithms that solve the well-known clustering problem. The procedure follows a simple and easy way to classify a given data set to a predefined, say K number of clusters. Determination of K is a difficult job and it is not known that which value of K can partition the objects as per our intuition. To overcome this problem we proposed K+ Means algorithm. This algorithm is an enhancement over K-Means algorithm.

## 1. Introduction

Hierarchical clustering[2][3], Partitional clustering[4], Bayesian clustering[5] are the different kind of clustering techniques. However, K-Means clustering algorithm is considered as the most popular clustering algorithm due to its simplicity and efficiency. This is the reason why we take K-Means algorithm for enhancement. We enhance the algorithm to K+ Means by taking the strengths of the algorithm and eliminated the weaknesses of the algorithm.

## 2. K-Means Algorithm

   *(a)* Place K points into the space represented by the objects that are being clustered. These points represent initial group centroids.
   *(b)* Assign each object to the group that has the closest centroid.
   *(c)* When all objects have been assigned, recalculate the positions of the K centroids.
   *(d)* Repeat Steps (b) and (c) until the centroids no longer move. This produces a separation of the objects into groups from which the metric to be minimized can be calculated.

The distance between objects are taken as Euclidian distance.

### 2.1. Strengths of K-Means Algorithm

The strengths of the algorithm is as follows:

   *(a)* Simple: easy to understand and to implement.
   *(b)* Efficient: Time complexity: $O(tkn)$, where n is the number of data points, k is the number of clusters, and t is the number of iterations. Since both k and t are small K-Means is considered a linear algorithm.

### 2.2. Weaknesses of K-Means Algorithm

The weaknesses of the algorithm is as follows:

   *(a)* The user needs to specify K.
   *(b)* The algorithm is sensitive to outliers. Outliers are data points that are very far away from other data points.

## 3. K+ Means Algorithm

   *(a)* Run K-Means algorithm to find K clusters.
   *(b)* Min, Max and Average intra cluster dissimilarity for each of the K clusters is computed.
   *(c)* The average intra cluster distance is expected to be almost similar and preferably small for each of the K clusters.
   *(d)* If for any cluster, the average value is more, its Max and Min values are checked. If the Max value is high, then outlier object is detected, as it has the maximum distance from its cluster representative.
   *(e)* Now, taking this outlier as another new representative, the algorithm is repeated to assign objects to K+1 cluster.
   *(f)* The algorithm is repeated until no more new representatives are formed and existing representatives do not change.

The distance between objects are taken as Euclidian distance.

## 3.1. Strengths of K+ Means Algorithm

The strengths of the algorithm is as follows:

(a) Simple: easy to understand and to implement.
(b) Efficient: Time complexity: $O(tk+^2n)$, where n is the number of data points, k+ is the number of clusters, and t is the number of iterations.
(c) The user only need to specify initial k value. Actual number of clusters will be obtained based on the data.
(d) The algorithm is not sensitive to outliers. In case of outliers, it will define a new cluster.

## 4. An Worked Out Example

Consider an example dataset shown in table-1 below:

| Object/Point | x | y |
|---|---|---|
| p1 | 1 | 4 |
| p2 | 1 | 3 |
| p3 | 2 | 2 |
| p4 | 7 | 2 |
| p5 | 8 | 3 |
| p6 | 9 | 2 |
| p7 | 5 | 6 |
| p8 | 6 | 7 |
| p9 | 7 | 6 |
| p10 | 8 | 7 |

Based on the above table the objects are plotted in 2D space as shown in figure-1 below:

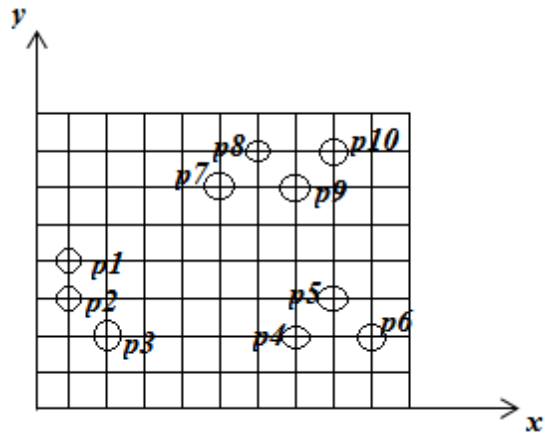

Figure-1: Example Data

## 4. 1. Application of K-Means Algorithm

Let us assume K=2 and select p1 and p5 as initial centroid. Now if we run K-means algorithm, we get Cluster 1 with {p1,p2,p3} and Cluster 2 with {P4,p5,p6,p7,p8,p9,p10}. The points (objects) and cluster centroids after running the K-means algorithm is shown in the figure-2 below:

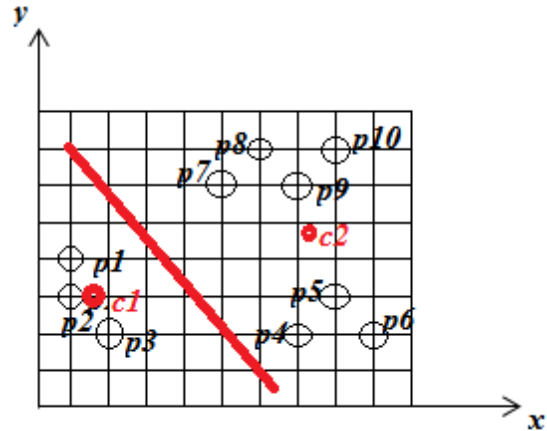

Figure-2: Clusters after running K-means algorithm

## 4. 2. Application of K+ Means Algorithm

Let us assume K=2 and select p1 and p5 as initial centroid. Based on that we get the initial clusters as shown in the figure-3 below:

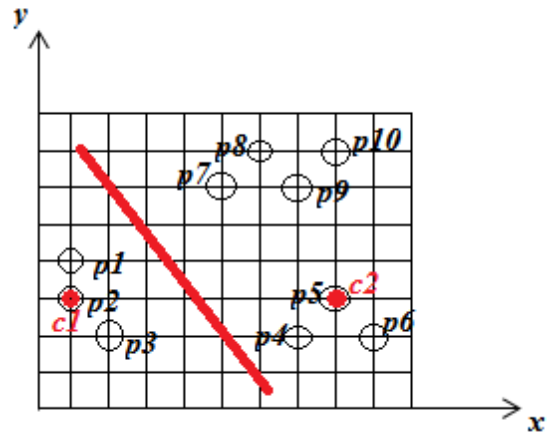

Figure-3: Initial clusters

Now we calculate the new centroids and get the clusters as shown in figure-4 below:

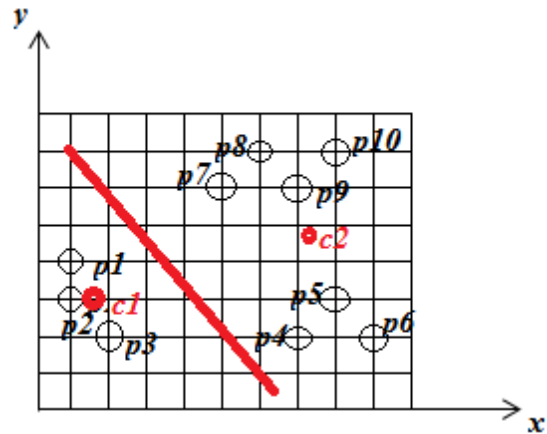

Figure-4: New cluster centroids

Now we calculate the intra-cluster distances of c1 and c2.
Consider c1: Min=0.33, Max=1.20, Avg=0.86
Consider c2: Min=1.30, Max=3.29 Avg=2.39

Here the avg value is comparatively high for c2. Max value is also high. So, we get the outlier object p6 as d(c2,p6)=3.29.

Taking c3(9,2) as new cluster centroid, we reassign the objects in c1, c2 and c3 as shown in figure-5 below:

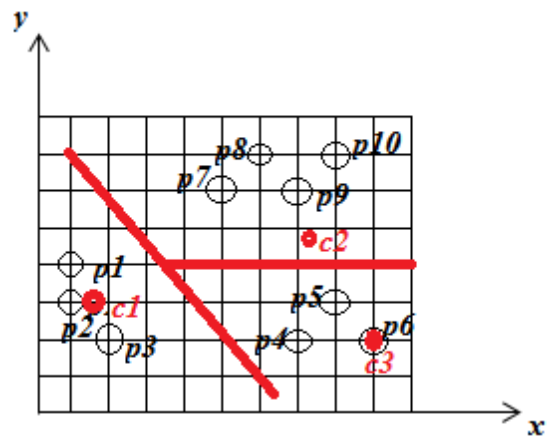

Figure-5: Initials centroids c1,c2,c3

Now we recalculate the centroids as shown in the figure-6 below:

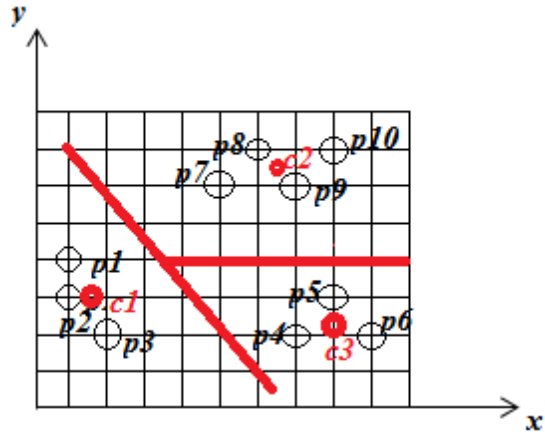

Figure-6: Modified centroids c1,c2,c3

Now we calculate the intra-cluster distances for c1, c2 and c3.
Consider c1: Min=0.33, Max=1.20, Avg=0.86
Consider c2: Min=0.71, Max=1.58, Avg=1.14
Consider c3: Min=0.67, Max=1.05, Avg=0.92
Here avg intra-cluster distances are similar for all three clusters, so we stop here.
So finally we get three clusters cluster 1={p1,p2,p3}, cluster2={p7,p8,p9,p10} and cluster3={p4,p5,p6}

### 4.3. Comparison:

K+ Means algorithm gives better clusters than K-Means algorithm. Naturally the complexity of K+ Means algorithm is little higher than K-Means algorithm. The below example shows how for outlier object two algorithms perform. Figure-7 shows how K-Means algorithm group objects when we run the algorithm taking K=2.

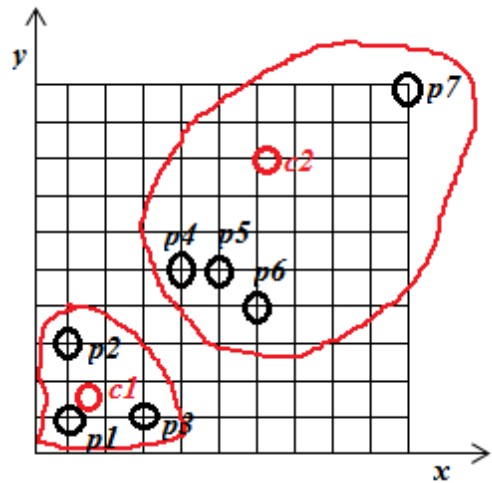

Figure-7: Grouping through K-Means algorithm

Figure-8 shows how K+ Means algorithm group objects when we run the algorithm taking K=2.

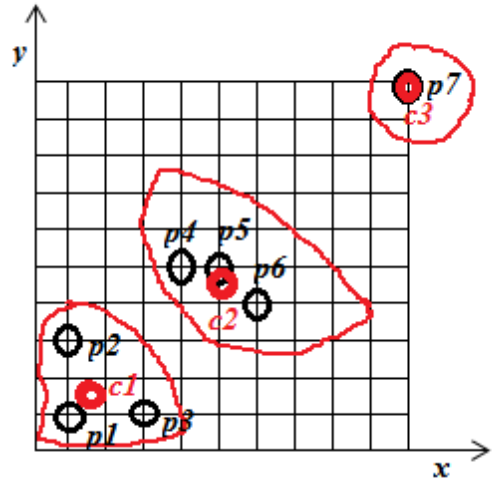

Figure-8: Grouping through K+ Means algorithm

## 9. Conclusion

In this paper we enhanced the K-Means algorithm to develop a new K+ Means algorithm. In this K+ Means algorithm all the strengths of K-Means algorithm is taken and the weaknesses of K-Means algorithms are removed.